\documentclass{article} 
\usepackage{iclr2027_conference,times}

\usepackage{amsmath,amsfonts,bm}

\def\eqref#1{equation~\ref{#1}}

\def\1{\bm{1}}

\DeclareMathAlphabet{\mathsfit}{\encodingdefault}{\sfdefault}{m}{sl}
\SetMathAlphabet{\mathsfit}{bold}{\encodingdefault}{\sfdefault}{bx}{n}

\usepackage{hyperref}
\usepackage{url}
\usepackage{graphicx}
\usepackage{subcaption}
\usepackage{wrapfig}
\usepackage{tcolorbox}
\usepackage{booktabs}
\tcbuselibrary{breakable}

\usepackage[utf8]{inputenc} 
\usepackage[T1]{fontenc}    
\usepackage{hyperref}       
\usepackage{url}            
\usepackage{booktabs}       
\usepackage{amsfonts}       
\usepackage{nicefrac}       
\usepackage{microtype}      
\usepackage{xcolor}         
\usepackage{amsmath}
\usepackage{algorithm}
\usepackage{algpseudocode}
\usepackage{amsmath, amssymb}

\title{A failure of Imagination: VLMs fail to use visual representations for text, but self-distillation can help}

\author{Shashwat Singh, He He, Tal Linzen \\
New York University \\
\texttt{\{ss20428,hhe,linzen\}@nyu.edu}
}

\iclrfinalcopy 
\begin{document}

\maketitle
\lhead{Preprint. Under review.}

\begin{abstract}
Vision-language (VL) models are typically built by fine-tuning a language model to attend to visual features in service of a Visual Question-Answering objective. We ask whether this alignment changes how language models represent \textit{text}---especially on tasks where visualization would help, like geometry questions posed in words. We distinguish \textit{weak} multimodality (the model can answer questions about an input image) from \textit{strong} multimodality (the model's text representations themselves become vision-aligned, improving performance on text-only tasks with latent visual structure).
To test for strong multimodality, we construct a task of shape-counting problems with paired equivalent text and image specifications. This task is trivial when the image is provided, but requires visualization when specified in text as line coordinates. We compare instruction-tuned Qwen-2.5 and Mistral models against their vision-aligned counterparts on the text variant of the task. We find that the multimodal models do no better in this setting, despite strong performance on the image version of the task. Probing experiments corroborate this: vision-aligned decoder embeddings carry no more visual structure than their text-only counterparts. Finally, we provide an optimization objective that distills in-context visual embeddings into text embeddings. This alignment induces gains on the text variant of the task without explicit task supervision, even on samples more complex than seen during alignment. This suggests that VQA-style alignment leaves available cross-modal transfer untapped.
\end{abstract}
\section{Introduction}
We study the paradigm of vision language alignment used in the Qwen \citep{bai2025qwen25vltechnicalreport, bai2025qwen3}, LlaVA \citep{liu2023visualinstructiontuning}, and Pixtral \citep{agrawal2024pixtral}. These models begin with a strong language decoder that is then aligned with a pretrained vision encoder on a visual question answering task. Since the visual task feedback is used to update the language backbone, we ask whether it teaches the language decoder to build meaningful visual representations of text when useful. 

We distinguish between \textit{weak multimodality} and \textit{strong multimodality}. A weakly multimodal model can operate on input from a different modality (vision) and can answer questions about that input. However, when it comes to input that is purely in text, a weakly multimodal model will behave exactly like the language-only model. By contrast, a strongly multimodal model can form representations for inputs provided in text that are vision-aligned and consequently perform better when the tasks are more naturally represented visually (for example, front-end code and edits, or solving geometry problems specified in words). Informally, strongly multimodal models treat all modalities as part of the same cognitive system rather than as separate input pipelines that share a decoder. We call this capability of a strongly multimodal model ``visualization'' in this work. We argue that strong multimodality is desirable because it would utilize data from a modality (say vision) to improve performance on another modality (text). 

To study visualization in a tractable way, we propose a synthetic task that has the following properties (i) For each instance, there is a way to equivalently specify the question in text or image; and (ii) The problem is much easier when specified visually. Our task consists of code that draws shapes and the language model is asked questions about the resulting image (see section~\ref{sec:data-setup}).
We evaluate Qwen 2.5 models \citep{bai2025qwen25vltechnicalreport} and Pixtral / Mistral Nemo 12B \citep{agrawal2024pixtral, mistral2024nemo}, which provide a clean minimal pair. 
We isolate the contribution of vision-language training to text-only reasoning by comparing vanilla text decoders to the corresponding VL-aligned text decoders. For both the model families, we find that vision alignment provides essentially no benefit on the text task, despite the image version being significantly easier for the same models. Linear and Contrastive probes on the decoder's hidden states confirm that the visual structure in both models' decoder representations are equally bad. 

Finally, we address whether visualization -- beyond what text training alone provides-- can be instilled at all. We propose an alignment method based on self-distillation: the model is trained to predict, from text, the embeddings it would have obtained given the image. Applied to our synthetic data (where paired samples are available) \textit{with no explicit task supervision}, this yields substantial improvements on the text version of the task across difficulty levels. This provides a proof of concept that alignment objectives aimed at strong multimodality, by targeting text representations directly, can yield gains on this task that VQA-style alignment does not.
\paragraph{Contributions. } Our contributions are three-fold: 
\begin{itemize}
    \item A new task that can be equivalently specified visually and in text, but is much easier to solve using visual representations. This allows us to evaluate the same models on the vision-specified version of the task and the text-specified version of the task while controlling for difficulty. Additionally, it also allows us to compare vision-aligned text decoders to pure text decoders with the upper-bound context of vision informed performance. 
    \item We present behavioral and representational evidence that for two widely used VL model families (Qwen 2.5 VL, Pixtral), the vision-aligned decoder does not construct visual representations for tasks provided in text, even when such representations are useful. 
    \item  To demonstrate that stronger multimodality can be instilled in multimodal models through better alignment methods, we provide a self distillation based method to vision-align a text student to learn to use visual representations for textual tasks. With this method, we provide a proof-of-concept model that operates on the text version of our task but with vision-informed representations -- demonstrating improvements across the difficulties of the task. 
\end{itemize}
\section{Related Work}
\subsection{Non-text Representations in Pure Language Models}
\cite{sharma2024vision}, \cite{patel2022mapping}, \cite{abdou2021can}, and \cite{ilharco2021probing} explore the question of whether visual grounding (similar to our notion of strong multimodality) emerges in vanilla language models. \cite{sharma2024vision} operationalize this in a code generation setting -- where they make the language models generate images from descriptions via code. \cite{ilharco2021probing} and \cite{abdou2021can} study whether language models learn visual relationships from their text training corpus -- \cite{ilharco2021probing} train mappings from BERT representations of concrete words to representations from a vision model (Faster R-CNN) and find nontrivial structural alignment while \cite{abdou2021can} do this by comparing embeddings to the CIELAB colour space. Finally, \cite{patel2022mapping} study grounded representations in language models (GPT-2 and GPT-3).  Our work asks a complementary question: what does vision-language alignment provide \textit{beyond} language-only training.
\cite{lepori2026languagemodelsstruggleuse} show that language models presented with a random walk over a latent state-space topology come to encode that topology in their hidden states (replicating \cite{park2025iclr}), yet largely fail to use those representations. 
\subsection{Visual Reasoning}
\cite{xu2026vision} and \cite{wang2025mathsight} both study vision-language models on geometry tasks. \cite{xu2026vision} investigate math problems that they can equivalently transcribe in images, text, and both together. They observe that the tested models (Qwen-3 \citep{bai2025qwen3}) simply do best on the text-only variants of the problems. \cite{wang2025mathsight} likewise report that the contribution of visual information diminishes on harder problems. These papers study whether visual information in-context can be effectively leveraged by vision-language models. By contrast, we study visual representations in purely textual input.
Our setup shares motivation with MentisOculi~\citep{zeller2026mentisoculi}, which designs tasks that require visual reasoning. They demonstrate that visual reasoning in models that can interleave text and image generation is limited by error accumulation. Our work differs in task specification: nearly all of their tasks include an image in the input, and on the one exception (RushHour) the models they test do \textit{worse} with visual context than without. Therefore, this setting does not allow us to answer our more specific question: \textit{In tasks where appropriate, does the vision-aligned text decoder perform better than a comparable text decoder with no vision training?}. Finally, \cite{xu2025visulogic} criticize existing vision-language benchmarks on the grounds that they can be solved from the text descriptions alone.

\subsection{Methods: Self-Distillation to Imagine Missing Modalities}
\cite{yang2025machine} provide a latent continuous reasoning methodology. For tasks where one can generate intermediate helper images for reasoning as training data, they propose training a multimodal model to generate the image encoding of the intermediate helper images instead. This is followed by finetuning on the task. Additionally, the idea of imitating a missing modality has been explored by \cite{li2024unified} and \cite{zheng2022boosting}.

\section{The Shape Counting Task}\label{sec:data-setup}
\begin{figure}[ht]
    \centering
    \begin{subfigure}[t]{0.70\textwidth}
        \centering
        \includegraphics[width=\linewidth]{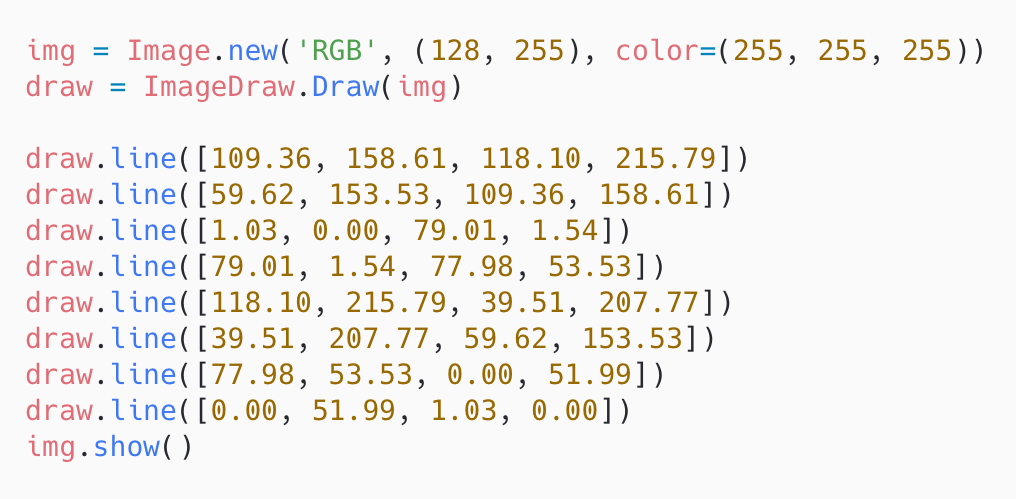}
        \caption{An example code snippet that would draw a configuration of a rectangle and a trapezium.}
        \label{fig:code-snippet}
    \end{subfigure}
    \hfill
    \begin{subfigure}[t]{0.20\textwidth}
        \centering
        \includegraphics[width=\linewidth]{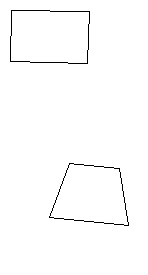}
        \caption{Output image with one trapezium and one rectangle. }
        \label{fig:demo-output}
    \end{subfigure}
    \caption{A simple example of the paired data sample: the code snippet (left) when compiled generates the image (right). The task is simply to count shapes of each kind given either the code snippet (for the text-variant of the task) or the rendered image (for the image variant).}
    \label{fig:code-snippet-illustration}
\end{figure}
We are concerned with problems that can be completely specified in text but would benefit a lot from intermediate computations in a visual representation space. To that end, we use a synthetic data setting for the experiments in this work. We describe it here briefly and provide the full data generation process in Appendix~\ref{app:data-generation}. We procedurally generate python code that when run would output a set of shapes (specifically, rectangles and trapeziums). This allows us to generate samples of varying difficulties that deterministically specify an image but do so entirely in text; for an illustrative example, see Figure~\ref{fig:code-snippet-illustration}. To generate an item, we sample the number of rectangles and trapeziums in the scene, rotate them randomly, and sample their positions in the image. We derive from the shapes a list of edges with exact coordinates that can be turned into a python program (see Figure~\ref{fig:code-snippet}). Crucially, we shuffle the order in which edges appear in the program. This leaves the rendered image unchanged but forces the model to compose edges that may be separated by long stretches of code.

\section{Vision-Language Models Use Poor Visual Representations}
\paragraph{Scope and Models.} This study examines a popular vision-language alignment paradigm, in which a pretrained language model is coupled to a vision encoder through a translation MLP block and trained under some alignment objective (usually visual question answering). To isolate the effect of alignment on the language model's representations, we require \textit{minimal pairs} of models, that is, the text decoder before visual adaptation and the updated weights after visual adaptation. These requirements leave us with the Qwen-VL family \citep{bai2025qwen25vltechnicalreport, bai2025qwen3}, Mistral / Pixtral \citep{mistral2024nemo, agrawal2024pixtral} and LLaVA \citep{liu2023visualinstructiontuning}. Furthermore, we need the VLM to do better on the image variant of the task than the text variant to conduct the study. This would validate that our synthetic data is in-distribution and the vision encoder has the features required for the image variant of the task. LLaVA fails this criterion but we present its results in Section~\ref{par:results_behavioral} for completeness. The analysis in this work centres around Qwen 2.5 and Pixtral models.
\subsection{Behavioral Experiments}\label{sec:behavioral-experiment}
The task is to count the number of shapes of each type given the source code. This requires the model to group edges into shapes and identify the angles in it. However, when the rendered image is available, the distinction (between rectangles and trapeziums) is obvious using heuristics in the image. 
We measure the performance of the model(s) on the task based on just the image; this establishes our upper-bound baseline. We then compare text-only models to vision-aligned text models on different answer modes (No-CoT, ICL-No-CoT, and ICL-CoT). In the \textsc{No-CoT} condition, the model is provided just with the description of the task, and is asked to output the answer immediately. In the \textsc{ICL-No-CoT} condition, the model is provided with the task description as well as examples of the task (i.e. in-context learning, or ICL), and is asked to output the answer immediately, Finally, in the \textsc{ICL-CoT} condition, the model is provided with examples and is allowed to output whatever it wants before providing the answer. For the image variant of the task, the model is simply asked to look at the image and count the number of shapes of each type (for the full prompts, see Appendix~\ref{app:prompts_used}). For all ICL setups, we use $3$ ICL examples. We calculate the accuracy of the task strictly: the model is given a point only if it gets the count of both the shapes right.
\begin{figure}[htbp]
    \centering
    \includegraphics[width=\textwidth]{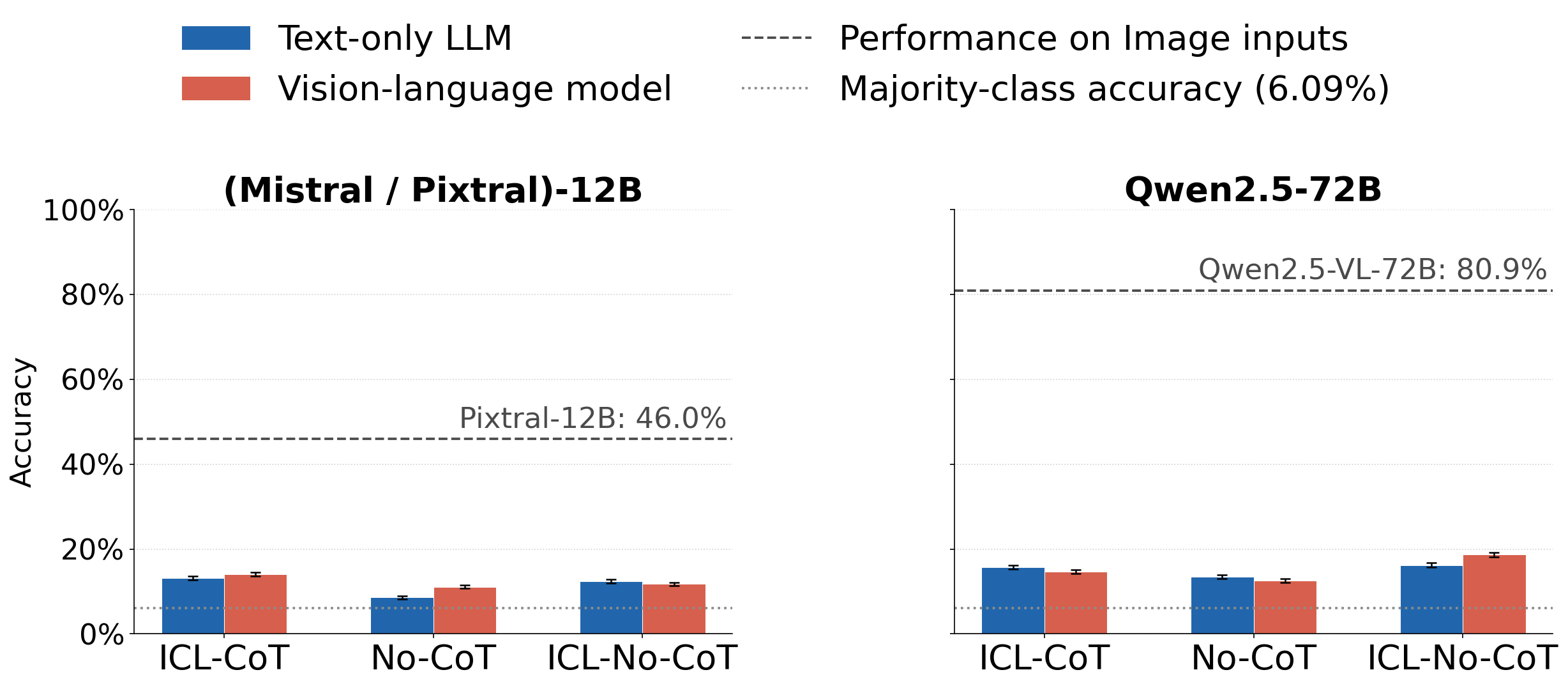}
    \caption{When the problem is provided in text, for all the settings tested---with and without in-context examples (ICL), and with and without chain of thought (CoT)---the performance of the multimodal models is similar to their text-only counterparts, and far below the  image ceiling. This indicates that vision alignment did not transfer visualization capabilities to the text decoder. Error bars are based on the 95\% confidence interval on accuracy estimates using the normal approximation to the binomial distribution.}
    \label{fig:behavioral}
\end{figure}
\paragraph{Results.}\label{par:results_behavioral} The results are shown in Figure~\ref{fig:behavioral}. When the task is provided visually, the models' performance sets a high ceiling. Neither of the text decoders come close to this ceiling on the text variant of the task. Most importantly, the vision-aligned decoder performs similarly to the text decoder. While ICL or COT don't seem to help our task, they improve the model's ability to simply count the total number of shapes (without identifying them). The models can often group vertices by looking at closed-loops during CoT (see Appendix~\ref{app:expt_1_copout}). Therefore, we restrict to rectangles and trapeziums as they have the same number of vertices and distinguishing them requires \textit{some} estimation of the angle. 
Additionally, we ran these experiments on LLaVA and the corresponding text backbone Llama-Vicuna \citep{zheng2023judgingllmasajudgemtbenchchatbot}. The vision-informed version of the task had an accuracy of 0.06 and the text variant performances were 0.07 for the multimodal decoder and 0.09 for the vanilla decoder. As the vision-informed performance is slightly worse, we exclude LLaVA from the remaining study. 
\subsection{Limited Evidence for Visual Representations in the Models' Embeddings}\label{sec:representation_expt}
The results in the shape-counting task suggest that vision-language aligned language decoders are not significantly better than their language-only counterparts. We next test if the internal representations of the vision-aligned models have more visual structure than those of the text-only models. We use both linear probes and contrastive alignment probes (for details, see Appendix~\ref{app:probe-details} and Appendix~\ref{app:contrastive-details}).
\paragraph{Linear Probes} The probe is trained to predict the number of times a particular type of shape (rectangle or trapezium) appears in the sample. This is a 9-class classification problem (an image can have $0-8$ rectangles). For each layer, we train separate probes on the mean-pooled representations for rectangle count and trapezium count. 
\paragraph{Contrastive Probes} Our dataset consists of paired image-text instances $(x_i, y_i)$ where  $x_i$ is a rendered image and $y_i$ is the code that produces it. We measure the alignment between the text model's embeddings of the code and the embeddings of the corresponding image produced by DiNOv2 \citep{oquab2023dinov2}, a pure vision encoder. We also conduct this experiment against a SigLip~\citep{zhai2023sigmoid} vision encoder.
We fit linear projections to the InfoNCE ~\citep{oord2018representation} (contrastive) loss popularized by CLIP~\citep{radford2021learning}. We refer to these as \emph{contrastive probes}. We extract a single vector for each: the CLS token from DINOv2 for $x_i$, and the mean-pooled hidden state at layer $\ell$ of the text model for $y_i$. Two linear projections $f_\theta : \mathbb{R}^{d_{\text{img}}} \to \mathbb{R}^{1536}$  and $f_\phi : \mathbb{R}^{d_{\text{txt}}} \to \mathbb{R}^{1536}$ map these into a shared space, where they are L2-normalized. The dimensionality of the shared space is that of DinoV2-Giant: 
$$
\tilde{z}_i^{\text{img}} = \tfrac{f_\theta(x_i)}{\lVert f_\theta(x_i) \rVert_2},
\qquad
\tilde{z}_i^{\text{txt}} = \tfrac{f_\phi(y_i)}{\lVert f_\phi(y_i) \rVert_2}.
$$
Given a batch of $N$ such pairs sampled uniformly from the dataset, we optimize the symmetric InfoNCE objective with temperature $\tau=0.5$:
\begin{equation}
\mathcal{L} = -\frac{1}{2N} \sum_{i=1}^{N} \left[
  \log \frac{\exp(\tilde{z}_i^{\text{img}} \cdot \tilde{z}_i^{\text{txt}} / \tau)}
            {\sum_{j=1}^{N} \exp(\tilde{z}_i^{\text{img}} \cdot \tilde{z}_j^{\text{txt}} / \tau)}
+ \log \frac{\exp(\tilde{z}_i^{\text{img}} \cdot \tilde{z}_i^{\text{txt}} / \tau)}
            {\sum_{j=1}^{N} \exp(\tilde{z}_j^{\text{img}} \cdot \tilde{z}_i^{\text{txt}} / \tau)}
\right].
\end{equation}
We report bidirectional top-1 batch retrieval accuracy: instance $i$ is correct in the image $\to$ text direction if $\arg\max_j \tilde{z}_i^{\text{img}} \cdot \tilde{z}_j^{\text{txt}} = i$, and analogously for text. We report the average of the two. Because both projections are linear, the retrieval accuracy can be treated as a proxy for the similarity in embedding structure between the two sets which we interpret as degree of visualization behavior. We argue that if the text model is doing computations that ``look like'' visualization at some layer then the representations at that level should have structure similar to the corresponding image embeddings from a good image encoder. We choose DiNo-v2 \citep{oquab2023dinov2} for the image embeddings as it is a \textit{pure} image encoder that was never aligned or supervised by text. We want to compare the degree of "visualization" between vision-aligned text models (Qwen-2.5VL, Pixtral) and their non-vision-aligned counterparts (Qwen-2.5, Mistral-Nemo). On our dataset, we know there is perfect correspondence between the text and the image data therefore text embeddings with visual structure should be easier to align with corresponding image embeddings. See Figure~\ref{fig:contrastive-probe} for results and Appendix~\ref{app:contrastive-details} for experimental details. To support these results we do the experiment with SigLip as the image encoder and see consistent results (see Appendix~\ref{app:contrastive_siglip}). Additionally, we also conduct this experiment on the COCO dataset~\citep{lin2015microsoftcococommonobjects}, a dataset with real images and free-form descriptive captions. The semantic correspondence of image-text pairs is noisier in COCO but the image data is more in-distribution for the image encoders used.  

\paragraph{Probe Controls} To rule out easy-to-exploit shortcuts in both probing setups we do a random model control. For the controls, all the probes are retrained on top of embeddings calculated by a randomly-initialized language decoder. Low performance on these control probes implies that the probes measure properties of the pretrained checkpoints beyond what can be inferred simply from surface-level statistics in the embeddings. 
\begin{figure}[t]
    \centering
    \includegraphics[width=\columnwidth]{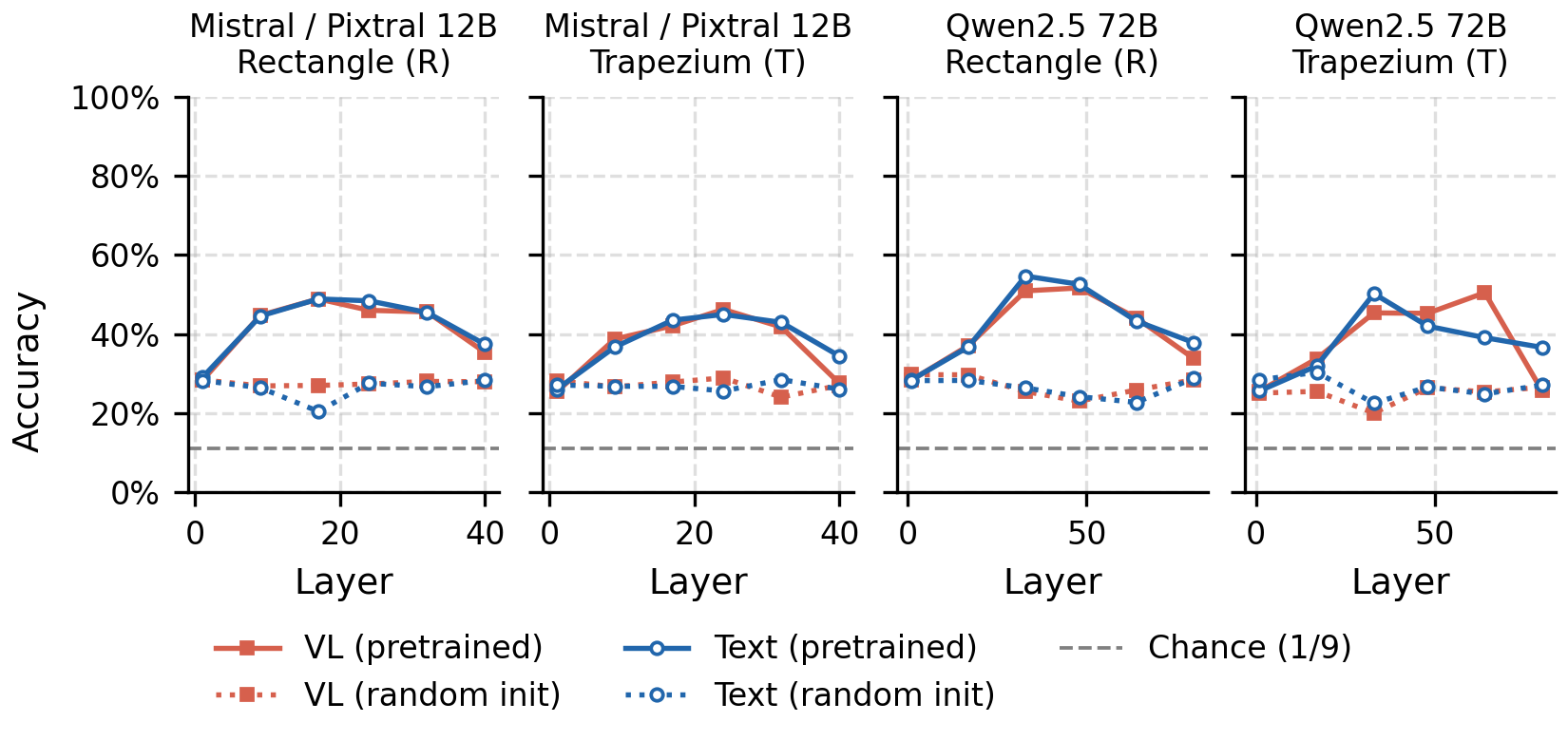}
    \caption{Vision-alignment does not add more visual structure to the embeddings of the multimodal model, compared to the text-only model. While probes trained on the embeddings from both classes of models show above-chance accuracy, the performance for the two classes of models is very similar.}
    \label{fig:linear-probe}
\end{figure}
\begin{figure}[h]
    \centering
    \includegraphics[width=\columnwidth]{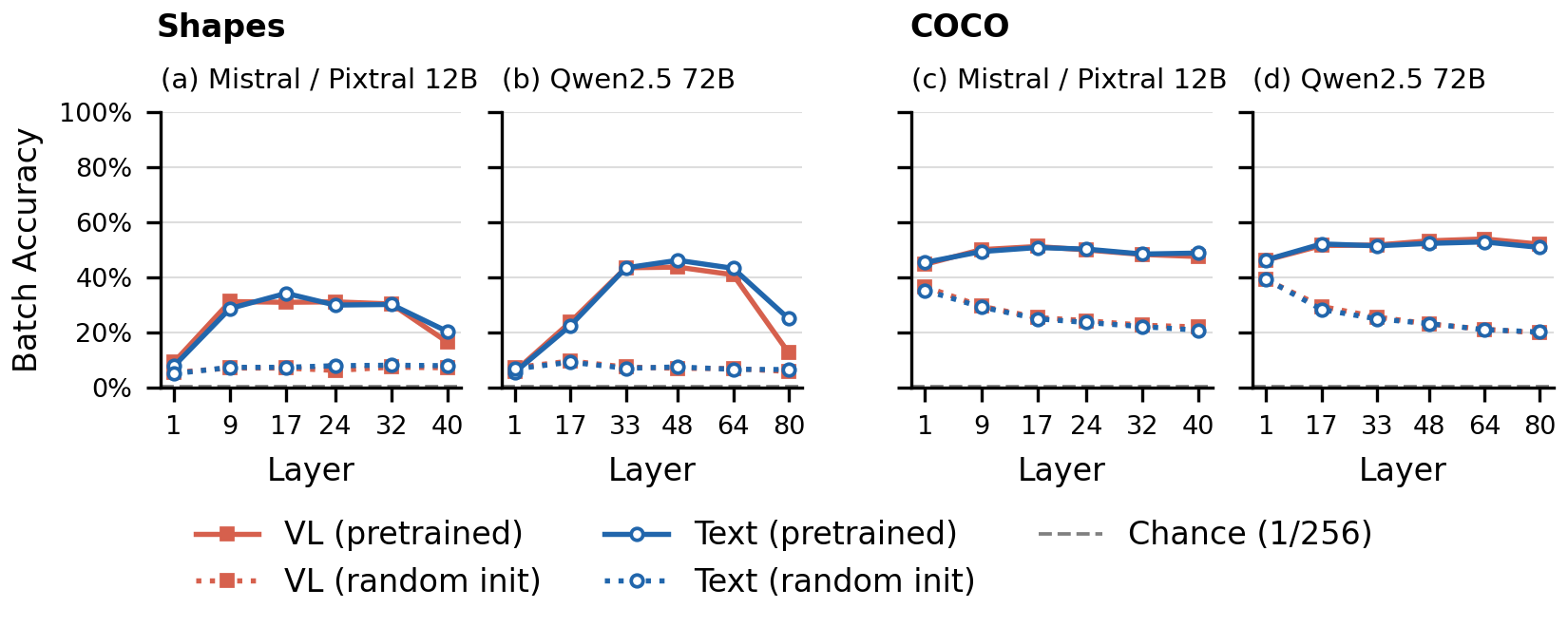}
    \caption{The trends on learnt retrieval corroborate that the multimodal decoder does not have better visual representations of textual inputs than the text-only model. Both the embedding sets get significantly above-chance and above-control retrieval accuracy against DiNo embeddings -- at least in the middle layers. However, the VL embeddings closely track text embeddings.}
    \label{fig:contrastive-probe}
\end{figure}
\paragraph{Results. } Both the linear probes (Figure~\ref{fig:linear-probe}) and the contrastive probes (Figure~\ref{fig:contrastive-probe}) show that, when the input is provided in text, the multimodal models typically do not encode visual structure better than their vanilla text counterparts. The contrastive probes on COCO show the same trend; this provides us with further confidence in our interpretation as the data in COCO more \textit{in-distribution} for the image encoders -- thereby ruling out the possibility that the results are an artifact of the data construction. We see that representational quality with respect to visualization peaks in the middle layers and then falls back towards the end, a finding consistent with \cite{skean2025layer}. We also note that all the probes significantly outperform the controls. These findings together with the behavioral findings suggest that, for these models, the VQA based vision-alignment paradigm does not add visual structure beyond what text training already provides. 
\section{Teaching the Model to Visualize}\label{sec:distil_section}
\begin{figure}[t]
  \centering
  \includegraphics[width=\textwidth]{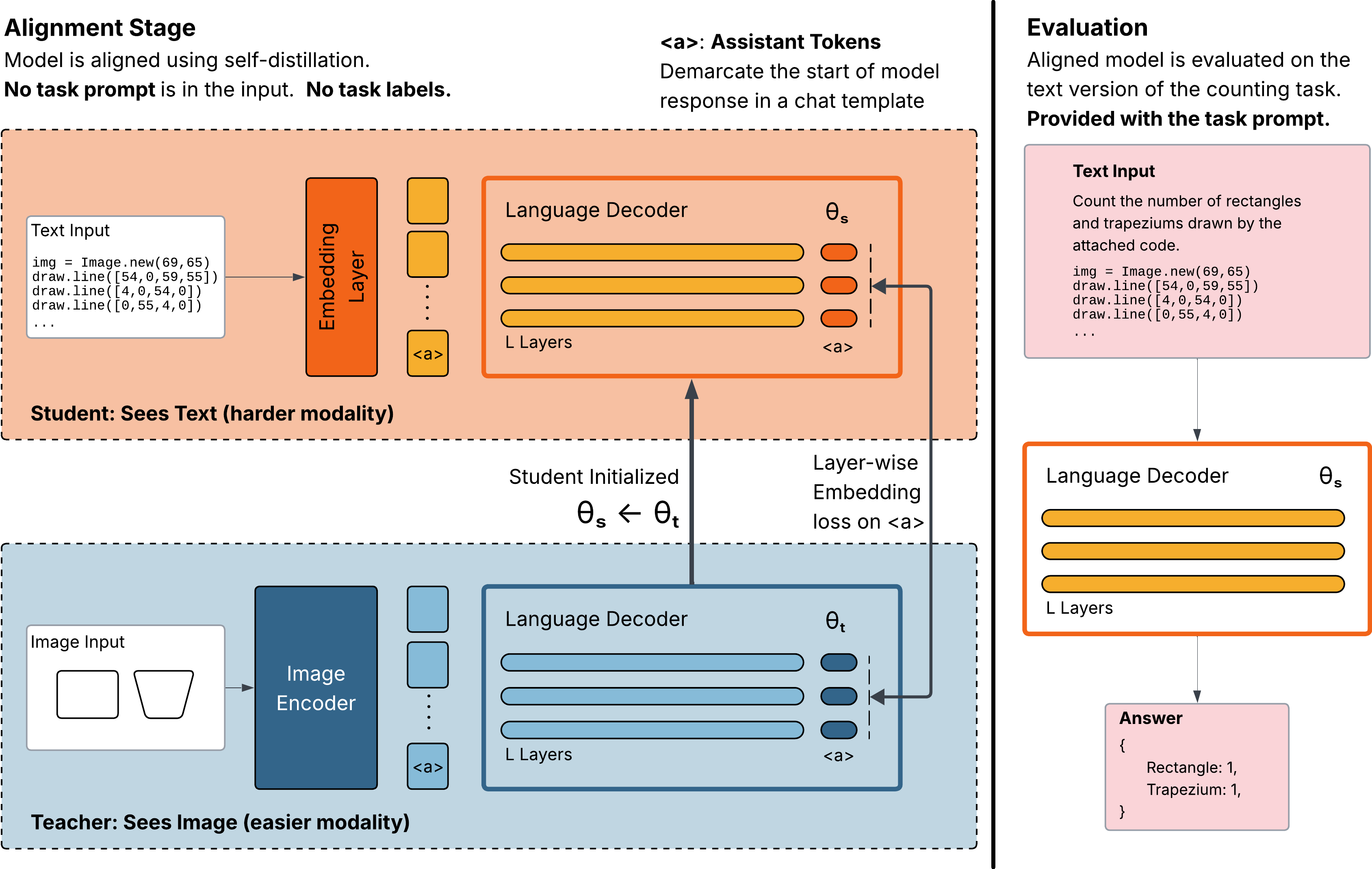}
  \caption{Given paired text-image data our alignment method trains the model to predict from text (code) the embeddings it would've had given the image. During alignment (left) we provide no task context or answer labels. During evaluation (right) the model performs the shape counting task.}
  \label{fig:distillation_demo}
\end{figure}

The problem, as identified, is that while vision-language models can pay attention to vision features to perform question-answering tasks (including the one we propose), they don't seem to construct internal representations that capture the visual structure of the problem. To encourage cross-modality transfer, we propose a methodology based on self-distillation. The intuition is that we will make the student model (initialized as a copy of the V+L teacher model) have the same intermediate embeddings \textit{without} the image as the teacher calculates \textit{with} the image. The Qwen 2.5 VL model treats vision features the same as token embeddings during language decoding i.e. vision patch embeddings are concatenated with text context (details about the Qwen architecture are in Appendix~\ref{app:qwen_arch}). This allows us to get meaningful transfer by literally equating the embeddings as opposed to some projection or KL-divergence loss which is typical in distillation literature. We illustrate this in Figure~\ref{fig:distillation_demo}.
\paragraph{Setup} Let $\mathcal{D} = \{(x_i, y_i)\}_{i=1}^{N}$ denote a dataset of paired image--text examples, where $x_i$ is an image and $y_i = (y_i^1, \dots, y_i^{T_i})$ is a tokenized text sequence drawn from a vocabulary $\mathcal{W}$. We assume access to a pretrained vision--language model (VLM) $f_\theta$ comprising a vision encoder $\Phi: \mathcal{X} \to \mathbb{R}^{M \times d}$ that maps an image to a sequence of $M$ visual embeddings, a text embedding layer $\mathcal{E}: \mathcal{W} \to \mathbb{R}^{d}$, and an $L$-layer transformer decoder $\mathcal{T}$ operating on sequences in $\mathbb{R}^{d}$, whose $\ell$-th layer produces hidden states in $\mathbb{R}^{d}$ for $\ell \in \{1, \dots, L\}$.
\paragraph{Teacher and Student} We initialize two copies of the pretrained $f_\theta$. The \emph{teacher} hereafter referred to as $f_{\theta _T}$ retains the full architecture and is kept frozen throughout training. For the \emph{student}, we discard the vision encoder; the student has to operate on text tokens only. Its parameters $\theta_S$ are trainable and initialized from $\theta$.
\paragraph{Input Construction} The teacher consumes the image-embeddings followed by the assistant token while the student consumes the text embeddings followed by the assistant token. We denote the assistant token as $\langle a \rangle$ 
\begin{equation}
    z_T^{(i)} \;=\; \Phi_{\theta _T}(x_i) \,\Vert\, \mathcal{E}_{\theta_T}(\langle a \rangle), \quad
    z_S^{(i)} \;=\; \mathcal{E}_{\theta_S}(y_i) \,\Vert\, \mathcal{E}_{\theta_S}(\langle a \rangle)
\end{equation}
Where $\Vert$ denotes sequence concatenation along the token axis and $\mathcal{E}$, the embedding layer, that applies to each token independently. 
\paragraph{Distillation objective.} For each example we perform a single forward pass of the teacher and student decoders on their respective input embedding sequences and extract the hidden states at the assistant generation position across all $L$ transformer layers:
\begin{align}
    h_T^{(i,\ell)} \;=\; \mathcal{T}^{(\ell)}_{\theta_T}\!\left(z_T^{(i)}\right)\!\big[-1\big], \quad
    h_S^{(i,\ell)} \;=\; \mathcal{T}^{(\ell)}_{\theta_S}\!\left(z_S^{(i)}\right)\!\big[-1\big],
\end{align}
for $\ell = 1, \dots, L$, where $\mathcal{T}^{(\ell)}(\cdot)[p]$ returns the $\ell$-th layer hidden state at position $p$ and we use $[-1]$ to indicate the last in the sequence (which refers to the representation of $\langle a \rangle$ token). The student is trained to match the teacher's representations at this position via the layerwise $\ell_2$ reconstruction loss. Note that for our experiments we try two variants of the reconstruction loss: vanilla MSE and normalized MSE; $\hat{h}$ in the following equation represents $h$ or $\frac{h}{\Vert h \Vert_2}$, respectively. We find that normalized MSE works better and report the corresponding results in Figure~\ref{fig:distil_results}.
\begin{equation}
    \mathcal{L}(\theta_S) \;=\; \mathbb{E}_{(x_i, y_i) \sim \mathcal{D}}
    \left[\, \frac{1}{L\,d} \sum_{\ell=1}^{L}
    \big\lVert \hat{h}_S^{(i,\ell)} \,-\, \operatorname{sg}\!\big(\hat{h}_T^{(i,\ell)}\big) \big\rVert_2^2
    \,\right],
\end{equation}
where $\operatorname{sg}(\cdot)$ denotes the stop-gradient operator, reflecting that the teacher is frozen. Note that the teacher conditions on $x_i$ via $\Phi(x_i)$ while the student conditions on $y_i$; the loss therefore forces $\theta_S$ to encode, from $y_i$, the internal state the multimodal teacher would have formed from $x_i$. 

\subsection{Experiment}
\begin{figure}[t]
    \centering
    \includegraphics[width=\textwidth]{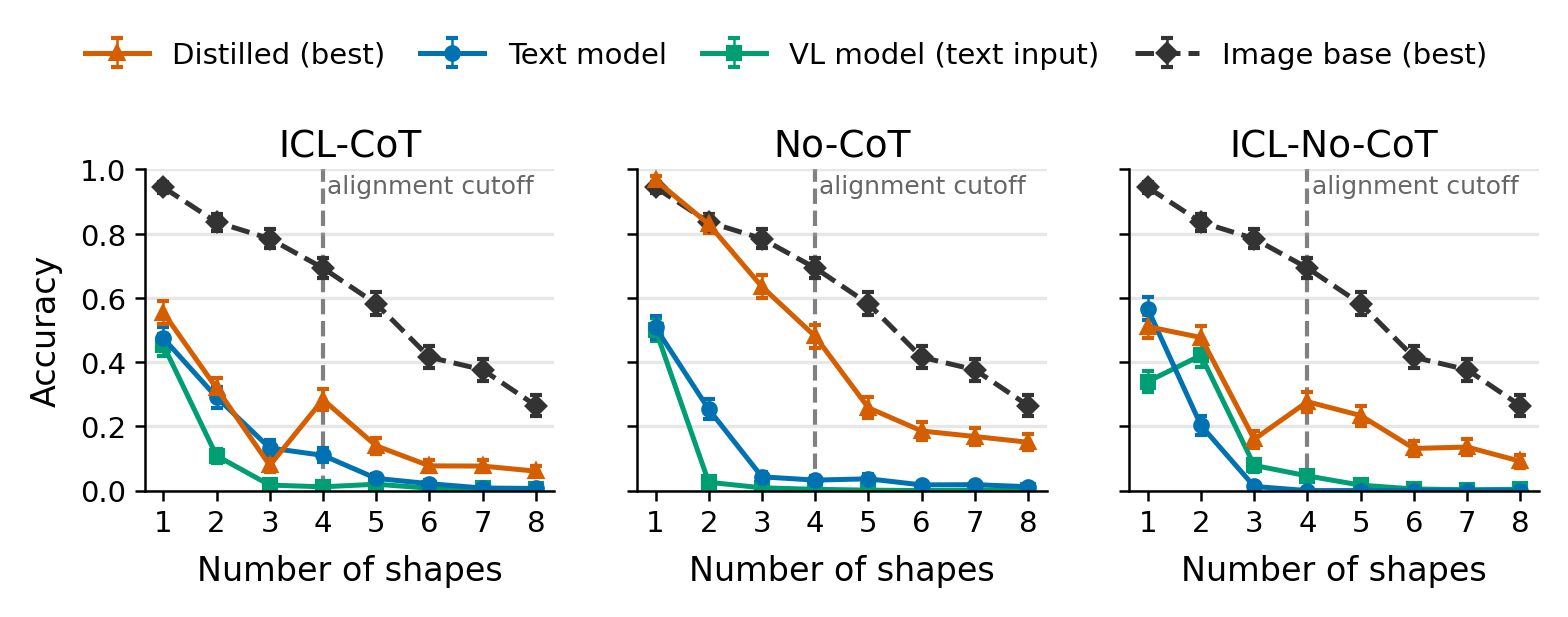}
    \caption{In all settings the distilled model outperforms the base models on across-shape accuracy even beyond the alignment cutoff, that is, the maximum number of shapes the distilled model has seen in a single sample. All the subplots represent the same set of samples -- just with different prompts. All the models in this experiment are 7B models (Qwen 2.5 Instruct and Qwen 2.5 VL Instruct).}
    \label{fig:distil_results}
\end{figure}
\paragraph{Alignment and Length Generalization} We experiment with a Qwen2.5-VL-7B-Instruct model; we use a 7B model, as opposed to the 72B or 12B models used earlier, because of compute constraints.
Crucially, during alignment, neither the student nor the teacher is provided with a specification of the task (an instruction to count the number of shapes) or task labels (the counts). Therefore, any transfer comes from the teacher's task-agnostic processing of the image. These image-conditioned representations are informative about properties such as shape counts, even though the teacher is never prompted to attend to them; this is precisely the kind of general visual structure we aim to transfer. To test for generalization, we only align the model on samples corresponding to a maximum of $4$ shapes per image (and code). Then, as in section~\ref{sec:behavioral-experiment}, we evaluate on a completely held-out set all the way up till $8$ shapes, this time with prompts and task specification. The experiment demonstrates: 1. Task agnostic alignment 2. Length generalization of alignment. The results are shown in Figure~\ref{fig:distil_results}. Note that all of them represent performance on the same set of samples and only the instruction prompts are different for the text inputs. The image performance curve is on the same instructions prompt for all the subplots. We provide the different instructions used in Appendix~\ref{app:prompts_used}. See Appendix~\ref{app:distillation_train_details} for other details. 

\paragraph{Results} Figure~\ref{fig:distil_results} shows that just the objective of predicting image-informed embeddings from the same model can show very strong gains on the text version of the task. The distillation model usually outperforms the vanilla text decoder and the multimodal decoder. 
It is important to note that the task was not specified during alignment -- this suggests that image encoder \textit{does} process the images in a generic way that is useful for the task \textit{post-facto} and hence learning to imitate \textit{that} computation allows our language model to visualize. The performance gains continue beyond the alignment cutoff, suggesting the stronger visual representations imparted by our distillation method generalize to samples that are harder to process than the ones used for alignment. We note that the gains in the ICL settings are modest, but this experiment is only intended as a proof-of-concept. 
\paragraph{Language Evaluations} We evaluate our distilled artifact on unrelated language benchmarks to check whether the distillation has catastrophic effects on language modeling. The distilled models show little to no degradation, See Appendix~\ref{app:lm-eval}.
\section{Discussion}
We test minimal pairs of language-only and vision-and-language models, on a task that can be equivalently specified in both text and image, but where the image version is substantially easier. We demonstrate that the standard vision-language alignment paradigm like in Qwen 2.5-VL and Pixtral do not make the language modeling backbone meaningfully better at text tasks that benefit from visualization. We conduct this analysis at two levels: behavioral and representational. Behaviorally, the vision-aligned text decoder and the vanilla text decoder perform comparably poorly on the text variant of our shape-counting task, lagging significantly behind the performance of the multimodal model when it is given the rendered image. Our representational probes show some evidence of visual structure emerging in the middle layers of both classes of models, but, crucially, the vision-aligned decoders' embeddings are typically on par with or worse than those of the vanilla text models. Together, these findings suggest that for models we study the VQA-based vision-alignment paradigm does not transfer vision priors into the language decoder's text-processing pathway: it only teaches the model to \textit{attend} to visual features when they are present, and not to \textit{think} visually about text.

A natural objection to this conclusion is that text training alone may already saturate whatever visual structure a language model can plausibly acquire. Prior work on emergent grounding in pure language models \citep{abdou2021can, sharma2024vision, patel2022mapping} has shown that nontrivial perceptual structure emerges in model that have only been trained on text. We address this objection with our distillation experiment. By aligning the Qwen-2.5-VL-7B text decoder so that, given a textual input, it reconstructs the hidden states the same model would produce given the corresponding image—without any task supervision—we induce measurable improvements on the text variant of our task. These gains length-generalize beyond the shape counts seen during alignment training. We interpret these results as a proof of concept for vision alignment that bleeds into the text pathways of a language model allowing it to use vision-informed representations for text tasks that benefit. 

We view our work as a call to review how vision-language alignment training is usually done i.e. visual alignment of a language model is mostly seen as a way for it to process images. Our results suggest that this does not, by itself, provide strong multimodality. If our goal is to get strong multimodality then alignment strategies that make the language model better at text tasks like the self distillation we demonstrate  deserve more attention 
\section{Limitations}
The results demonstrated in this work including the proposed alignment methodology has been conducted on a single synthetic setup. Future work should expand the scope to other evaluative setups where the image-based performance is significantly better than the equivalently specified text-based performance cleanly. Furthermore, the distillation results are provided as a proof of concept. We leave adapting to scale of data for production-ready models to future work. 

\section{AI Disclosure}
LLMs were used to help code and manage the experiments reported in this paper i.e. for implementing methods. The written code was verified by the authors for correctness. They were also used to polish the document, including the figures and writing. We take responsibility for the final content of this work, including text, claims, or artifacts produced with the aid of generative AI.

\bibliography{iclr2027_conference}
\bibliographystyle{iclr2027_conference}

\appendix
\section{Appendix}

\subsection{Data Generation Protocol}\label{app:data-generation}

Each sample consists of a small scene of non-overlapping shapes placed along a 
random walk on a 2D plane, together with a shuffled list of the edge lengths 
of those shapes. The generation procedure is summarized in 
Algorithm~\ref{alg:sample}; we describe its components here.

\paragraph{Shapes and directions.}
We draw shapes from the set $\mathcal{S} = \{\texttt{rectangle}, \texttt{trapezium}\}$. 
A rectangle is parameterized by two side lengths $(a, b)$, and an isosceles 
trapezium by its top length $t$, bottom length $b$, and height $h$. For each 
shape we deterministically construct its vertices at the origin via the helpers 
$\textsc{RectVertices}$ and $\textsc{TrapVertices}$, and connect them into a 
closed polygon with $\textsc{CyclicEdges}$.

Placements are connected by unit steps in one of four cardinal directions, 
$\mathcal{D} = \{\texttt{up}, \texttt{down}, \texttt{left}, \texttt{right}\}$, 
with associated unit vectors $\{\mathbf{e}_d\}_{d \in \mathcal{D}}$. Successive 
shapes are placed at positions $p_1, p_2, \ldots$ where 
$p_{i+1} = p_i + \rho\, \mathbf{e}_{c_i}$ and $\rho > 0$ controls the spacing.

\paragraph{Sampling procedure.}
Given a sample, we first draw the number of shapes $n \sim \mathrm{Uniform}\{1, \ldots, N_{\max}\}$, 
their types $s_1, \ldots, s_n \stackrel{\text{i.i.d.}}{\sim} \mathrm{Uniform}(\mathcal{S})$, 
and the step directions $c_1, \ldots, c_{n-1} \stackrel{\text{i.i.d.}}{\sim} \mathrm{Uniform}(\mathcal{D})$ 
between successive placements. For each shape $i$ we draw its dimensions 
uniformly from $[\ell_{\min}, \ell_{\max}]$ (and additionally $h_i \in [h_{\min}, h_{\max}]$ 
for trapezia), construct its vertices at the origin, and apply a rotation by 
$\theta_i \sim \mathrm{Uniform}(\Theta)$ followed by a translation to $p_i$. 
Letting $R(\theta)$ denote the $2 \times 2$ rotation matrix, each edge $(u, v)$ 
of shape $i$ contributes the rotated, translated edge 
$\big(R(\theta_i) u + p_i,\, R(\theta_i) v + p_i\big)$ to the global edge set $E$.

\paragraph{Output.}
Given the final edge set $E$, we sample a random permutation $\sigma$ of its 
elements and return the sequence of edge lengths $\mathcal{P} = (\ell(e_{\sigma(1)}), \ldots, \ell(e_{\sigma(|E|)}))$ 
together with a rendered image of the scene. The shuffling forces the text version to require the model to compose edges separated by other lines of code. 

\begin{algorithm}[H]
\caption{Sample generation}
\label{alg:sample}
\begin{algorithmic}[1]
\Require $N_{\max}$, $[\ell_{\min}, \ell_{\max}]$, $[h_{\min}, h_{\max}]$, $\rho$, $\Theta$
\State $n \sim \mathrm{Uniform}\{1, \ldots, N_{\max}\}$
\State $s_i \stackrel{\text{i.i.d.}}{\sim} \mathrm{Uniform}(\mathcal{S})$ for $i = 1, \ldots, n$
\State $c_i \stackrel{\text{i.i.d.}}{\sim} \mathrm{Uniform}(\mathcal{D})$ for $i = 1, \ldots, n-1$
\State $p_1 \gets \mathbf{0}$;\quad $E \gets \emptyset$
\For{$i = 1, \ldots, n$}
    \State $\theta_i \sim \mathrm{Uniform}(\Theta)$
    \If{$s_i = \texttt{rectangle}$}
        \State $(a_i, b_i) \stackrel{\text{i.i.d.}}{\sim} \mathrm{Uniform}([\ell_{\min}, \ell_{\max}])^2$
        \State $V_i \gets \textsc{RectVertices}(a_i, b_i)$
    \Else \Comment{$s_i = \texttt{trapezium}$}
        \State $(t_i, b_i) \stackrel{\text{i.i.d.}}{\sim} \mathrm{Uniform}([\ell_{\min}, \ell_{\max}])^2$
        \State $h_i \sim \mathrm{Uniform}([h_{\min}, h_{\max}])$
        \State $V_i \gets \textsc{TrapVertices}(t_i, b_i, h_i)$
    \EndIf
    \State $E_i \gets \textsc{CyclicEdges}(V_i)$
    \State $E \gets E \cup \big\{ \big(R(\theta_i)\, u + p_i,\ R(\theta_i)\, v + p_i\big) : (u, v) \in E_i \big\}$
    \If{$i < n$}
        \State $p_{i+1} \gets p_i + \rho \cdot \mathbf{e}_{c_i}$
    \EndIf
\EndFor
\State $\sigma \sim \mathrm{Uniform}(\mathrm{Permutations}(E))$
\State $\mathcal{P} \gets \big(e_{\sigma(1)}, \ldots, e_{\sigma(|E|)}\big)$
\State \Return $(\mathcal{P}, \textsc{Render}(\mathcal{P}))$
\end{algorithmic}
\end{algorithm}

\subsection{Different Prompts Used}\label{app:prompts_used}

\subsubsection{ICL + COT}\label{app:prompt_icl_cot}

This is the prompt we use for the setting where the model is provided with ICL samples and is allowed to do chain of thought. 

\begin{tcolorbox}[breakable, colback=gray!10!white, colframe=black!75!white, title=User Prompt]
\begingroup\obeylines
Your task is to analyze the snippet of code in detail and identify the shapes it draws.

Follow these steps:
1. Parse coordinates: List each draw.line([x1, y1, x2, y2]) with start and end points.
2. Group connected lines: Identify which lines connect to form closed polygons.
3. Classify the shape: Decide if it is a rectangle, trapezium, triangle, or another polygon.
4. Locate the shape: Describe its approximate position (e.g., "top-left", "center-right").
5. Visual reasoning: Consider how the shapes would look on the canvas.
6. Count shapes: Report how many of each type of shape exist.
7. Summarize:
   - Give a natural-language description of all the shapes and their layout.
   - Provide a dictionary that lists the shape names and how many times each occurs.

Format your final answer like this:

Analysis:
<your reasoning and description here>

Shapes:
{
  "Rectangle": <count>,
  "Trapezium": <count>,
}

Important note: You will be given images that have either rectangles or trapeziums. No other shape. 

Here are few examples with their right answers:

Figure 1:

```
img = Image.new('RGB', (69, 65), color=(255, 255, 255))
draw = ImageDraw.Draw(img)
draw.line([54.50, 0.00, 59.00, 55.00], fill='black', width=1)
draw.line([4.50, 0.00, 54.50, 0.00], fill='black', width=1)
draw.line([0.00, 55.00, 4.50, 0.00], fill='black', width=1)
draw.line([59.00, 55.00, 0.00, 55.00], fill='black', width=1)
img.show()
```

The image generated by the above code has the following shapes: 

Shapes:
{
  "Trapezium": 1,
}

Figure 2:
```
img = Image.new('RGB', (75, 85), color=(255, 255, 255))
draw = ImageDraw.Draw(img)
draw.line([3.25, 0.00, 65.19, 2.76], fill='black', width=1)
draw.line([61.94, 75.69, 0.00, 72.93], fill='black', width=1)
draw.line([65.19, 2.76, 61.94, 75.69], fill='black', width=1)
draw.line([0.00, 72.93, 3.25, 0.00], fill='black', width=1)
img.show()
```

The image generated by the above code has the following shapes: 

Shapes:
{
  "Rectangle": 1,
}

Figure 3:
```
img = Image.new('RGB', (147, 257), color=(255, 255, 255))
draw = ImageDraw.Draw(img)
draw.line([137.00, 247.00, 44.00, 247.00], fill='black', width=1)
draw.line([88.00, 0.00, 88.00, 90.00], fill='black', width=1)
draw.line([88.00, 90.00, 0.00, 90.00], fill='black', width=1)
draw.line([115.50, 190.00, 137.00, 247.00], fill='black', width=1)
draw.line([0.00, 90.00, 0.00, 0.00], fill='black', width=1)
draw.line([44.00, 247.00, 65.50, 190.00], fill='black', width=1)
draw.line([0.00, 0.00, 88.00, 0.00], fill='black', width=1)
draw.line([65.50, 190.00, 115.50, 190.00], fill='black', width=1)
img.show()
```

The image generated by the above code has the following shapes: 

Shapes:
{
  "Rectangle": 1,
  "Trapezium": 1,
}

Figure 4:
```
img = Image.new('RGB', (128, 225), color=(255, 255, 255))
draw = ImageDraw.Draw(img)
draw.line([109.36, 158.61, 118.10, 215.79], fill='black', width=1)
draw.line([59.62, 153.53, 109.36, 158.61], fill='black', width=1)
draw.line([1.03, 0.00, 79.01, 1.54], fill='black', width=1)
draw.line([79.01, 1.54, 77.98, 53.53], fill='black', width=1)
draw.line([118.10, 215.79, 39.51, 207.77], fill='black', width=1)
draw.line([39.51, 207.77, 59.62, 153.53], fill='black', width=1)
draw.line([77.98, 53.53, 0.00, 51.99], fill='black', width=1)
draw.line([0.00, 51.99, 1.03, 0.00], fill='black', width=1)
img.show()
```

The image generated by the above code has the following shapes: 

Shapes:
{
  "Rectangle": 1,
  "Trapezium": 1,
}

Now, Do whatever analysis you need to understand the shapes drawn by the code snippet below:
\endgroup
\end{tcolorbox}

\subsubsection{ICL + No COT}\label{app:prompt_icl_no_cot}

This is the prompt for the setting where the model is provided with ICL examples and is asked to not do chain of thought. 

\begin{tcolorbox}[breakable, colback=gray!10!white, colframe=black!75!white, title=User Prompt]
\begingroup\obeylines

Your task is to analyze the snippet of code in detail and identify the shapes it draws. Specifically, you need to count the number of shapes of each kind. 

The answer should be in the form of a dictionary with shape names and counts.

Important note: You will be given images that have either rectangles or trapeziums. No other shape. 

Here are few examples with their right answers:

Figure 1:

```
img = Image.new('RGB', (69, 65), color=(255, 255, 255))
draw = ImageDraw.Draw(img)
draw.line([54.50, 0.00, 59.00, 55.00], fill='black', width=1)
draw.line([4.50, 0.00, 54.50, 0.00], fill='black', width=1)
draw.line([0.00, 55.00, 4.50, 0.00], fill='black', width=1)
draw.line([59.00, 55.00, 0.00, 55.00], fill='black', width=1)
img.show()
```

The image generated by the above code has the following shapes: 

Shapes:
{
  "Trapezium": 1,
}

Figure 2:
```
img = Image.new('RGB', (75, 85), color=(255, 255, 255))
draw = ImageDraw.Draw(img)
draw.line([3.25, 0.00, 65.19, 2.76], fill='black', width=1)
draw.line([61.94, 75.69, 0.00, 72.93], fill='black', width=1)
draw.line([65.19, 2.76, 61.94, 75.69], fill='black', width=1)
draw.line([0.00, 72.93, 3.25, 0.00], fill='black', width=1)
img.show()
```

The image generated by the above code has the following shapes: 

Shapes:
{
  "Rectangle": 1,
}

Figure 3:
```
img = Image.new('RGB', (147, 257), color=(255, 255, 255))
draw = ImageDraw.Draw(img)
draw.line([137.00, 247.00, 44.00, 247.00], fill='black', width=1)
draw.line([88.00, 0.00, 88.00, 90.00], fill='black', width=1)
draw.line([88.00, 90.00, 0.00, 90.00], fill='black', width=1)
draw.line([115.50, 190.00, 137.00, 247.00], fill='black', width=1)
draw.line([0.00, 90.00, 0.00, 0.00], fill='black', width=1)
draw.line([44.00, 247.00, 65.50, 190.00], fill='black', width=1)
draw.line([0.00, 0.00, 88.00, 0.00], fill='black', width=1)
draw.line([65.50, 190.00, 115.50, 190.00], fill='black', width=1)
img.show()
```

The image generated by the above code has the following shapes: 

Shapes:
{
  "Rectangle": 1,
  "Trapezium": 1,
}

Figure 4:
```
img = Image.new('RGB', (128, 225), color=(255, 255, 255))
draw = ImageDraw.Draw(img)
draw.line([109.36, 158.61, 118.10, 215.79], fill='black', width=1)
draw.line([59.62, 153.53, 109.36, 158.61], fill='black', width=1)
draw.line([1.03, 0.00, 79.01, 1.54], fill='black', width=1)
draw.line([79.01, 1.54, 77.98, 53.53], fill='black', width=1)
draw.line([118.10, 215.79, 39.51, 207.77], fill='black', width=1)
draw.line([39.51, 207.77, 59.62, 153.53], fill='black', width=1)
draw.line([77.98, 53.53, 0.00, 51.99], fill='black', width=1)
draw.line([0.00, 51.99, 1.03, 0.00], fill='black', width=1)
img.show()
```

The image generated by the above code has the following shapes: 

Shapes:
{
  "Rectangle": 1,
  "Trapezium": 1,
}

Respond with only the answer dictionary for the code snippet below:

\endgroup
\end{tcolorbox}

\subsubsection{Prompt with only the task description}\label{app:prompt_task_description}

This is the prompt where the prompt is only provided with the task description. 

\begin{tcolorbox}[breakable, colback=gray!10!white, colframe=black!75!white, title=User Prompt]
\begingroup\obeylines

Your task is to analyze the snippet of code in detail and identify the shapes it draws. Sometimes you will have the image it draws.

Specifically, you need to count the number of shapes of each kind. 

The answer should be in the form of a dictionary with shape names and counts.

Important note: You will be given images that have either rectangles or trapeziums. No other shape. 

Stick to this output format: 

{
    "Rectangle": <count>,
    "Trapezium": <count>
}

\endgroup
\end{tcolorbox}

\subsubsection{Prompt for image input}\label{app:prompt_image_input}

\begin{tcolorbox}[breakable, colback=gray!10!white, colframe=black!75!white, title=User Prompt]
\begingroup\obeylines
Your task is to analyse the image and count the number of shapes of each kind in it. 

The answer should be in the form of a dictionary with shape names and counts. 

Important note: You will be given images that have rectangles and / or trapeziums. No other shape. 

Look at each shape in this image and analyse whether it is a rectangle or a trapezium and tell me the counts. 

Format your answer as a dictionary like so: 

{
    'Rectangle': <count>,
    'Trapezium': <count>, 
}

\endgroup
\end{tcolorbox}

\subsection{Summarising the Qwen 2.5 Architecture}\label{app:qwen_arch}

We provide a summary description of the Qwen 2.5 VL models here to help with the intuition behind some design decisions for the distillation objective specified in Section~\ref{sec:distil_section}.

The Qwen 2.5 VL models \citep{bai2025qwen25vltechnicalreport} consist of three main components (i) the vision encoder which is a ViT with some efficiency modifications (ii) an MLP-based vision-language merger component (iii) Large Language Model which starts off as the Qwen 2.5 LM. An inference forward pass goes through the following: the image is processed by the ViT and the corresponding features are passed through the MLP merger block, following this the patch vectors are simply put in the context of the language model like the embedding of any other token. The Language Model decoder attends to the image information and the text information auto-regressively. 

The vision encoder is pre-trained using a CLIP-style objective first. In the second-stage, all the components including the vision encoder, the MLP-merger, and the language model are kept un-frozen and trained for a visual question answering task. This is followed by a third stage to adapt the model to work with long-context videos. 



\subsection{Experiment Hyper-parameters}\label{app:training_details}

\subsubsection{Linear-Probe Layer Sweep Details}
\label{app:probe-details}

Hyperparameters for the linear probe from Section~\ref{sec:representation_expt}.

\begin{itemize}
  \item \textbf{Layers swept:} We conduct the probing experiments on evenly spaced 6 layers for each of the models. 
  \item \textbf{Inference:} bfloat16 with FlashAttention-2 under
        HuggingFace tensor parallelism (\texttt{torchrun}, one process
        per GPU, all visible GPUs).
  \item \textbf{Pooling:} mean-pooling.
  \item \textbf{Optimizer:} Adam, learning rate $10^{-3}$ (PyTorch
        defaults otherwise); no weight decay, no gradient clipping.
  \item \textbf{Schedule:} at most 100 epochs with early stopping on
        validation loss (patience 10).
  \item \textbf{Batch size:} 32 (train/val/test); training batches are
        \emph{not} shuffled.
  \item \textbf{Precision / hardware:} probe trains in fp32 on a single
        GPU.
  \item \textbf{Data:} $12{,}000$ synthetic scenes, at most 8 shapes,
        sizes in $[20, 100]$ px, random rotation, $10$\,px padding,
        shuffled description lines; sequential (not random) 80/10/10
        train/validation/test split. Text embeddings only --- no images
        are used.
  \item \textbf{Metrics:} macro-averaged accuracy, precision, and recall
        per head.
\end{itemize}

\subsubsection{Contrastive Text--Image Probe Details}
\label{app:contrastive-details}

Hyperparameters for the contrastive probe from Section~\ref{sec:representation_expt}.

\begin{itemize}
  \item \textbf{Text embeddings:} extracted exactly as in
        Appendix~\ref{app:probe-details} (same models, same six layers,
        same random re-initialization, masked-mean pooling, bfloat16
        tensor-parallel inference).
  \item \textbf{Image embeddings:} frozen DINOv2-giant, final-layer CLS
        token (dimension $1536$); the image-side layer is \emph{not}
        swept. Images are rendered from the same scene descriptions, so
        each sample yields an aligned (text, image) pair. For SigLIP, the image embedding and shared projection dimension is 1152.
  \item \textbf{Probe:} one linear projection per modality; shared projection
        dimension $=\min(d_\text{text}, d_\text{image})$), with no
        hidden layers or activations. 
  \item \textbf{Loss:} symmetric InfoNCE on $\ell_2$-normalized
        projections --- cross-entropy over the in-batch similarity
        matrix, averaged over the text$\to$image and image$\to$text
        directions --- with a fixed temperature of $0.5$.
  \item \textbf{Optimizer:} Adam, learning rate $10^{-3}$; no early
        stopping, no gradient clipping.
  \item \textbf{Schedule:} 10 epochs.
  \item \textbf{Batch size:} 256; batches drawn by a repeated-shuffle
        sampler with sample multiplier 1.5 (each ``epoch'' covers 1.5
        shuffled passes; no sample appears twice within a batch). The
        same sampler is used for validation and test, so evaluation
        metrics average over $\sim$1.5 shuffled passes of the split.
  \item \textbf{Precision / hardware:} probe trains in fp32 on a single
        GPU.
  \item \textbf{Data:} same generation settings and sequential 80/10/10
        split as Appendix~\ref{app:probe-details} ($12{,}000$ samples,
        $9{,}600$ train).
  \item \textbf{Evaluation:} in-batch retrieval accuracy --- within each
        batch of 256 pairs, the fraction of samples whose highest cosine
        similarity is their true counterpart, averaged over both
        retrieval directions (chance $= 1/256$).
\end{itemize}

For the COCO variant of the contrastive probe experiment, everything is identical except the dataset. We use the train / val / test splits from the actual dataset.

\subsubsection{Distillation Training Details}\label{app:distillation_train_details}

Hyperparameters for the distillation experiment from Section~\ref{sec:distil_section}.

\begin{itemize}
  \item Optimizer: AdamW, learning rate $5\times10^{-5}$
        (selected from a sweep over $\{10^{-6}, 5\times10^{-6}, 10^{-5},
        5\times10^{-5}\}$); PyTorch defaults otherwise
        ($\beta=(0.9, 0.999)$, $\epsilon=10^{-8}$, weight decay $0.01$).
  \item Schedule: cosine decay with linear warmup over the first
        $1{,}000$ steps, decaying over the estimated total number of
        training steps; stepped per optimizer step.
  \item Gradient clipping: global norm $1.0$ (computed collectively
        across FSDP shards).
  \item Epochs: We early-stop based on performance on a held-out validation set. The reported results are for the 10th epoch of a 40 epoch run. 
  \item Batch size: 4
  \item Precision: full fp32 (no mixed precision).
\end{itemize}

The best model across sweeps was picked based on performance on a held-out validation set of 2000 samples. The test set consists of $6{,}000$ held-out samples. 

\subsection{Contrastive Probes With a SigLip as an Image Encoder}\label{app:contrastive_siglip}

We support the results of the contrastive probes in Section~\ref{sec:representation_expt} by rerunning them with a different image encoder as the ``image reference''. Specifically, we use SigLIP~\citep{zhai2023sigmoid}.

\begin{figure}[h]
    \centering
    \includegraphics[width=\columnwidth]{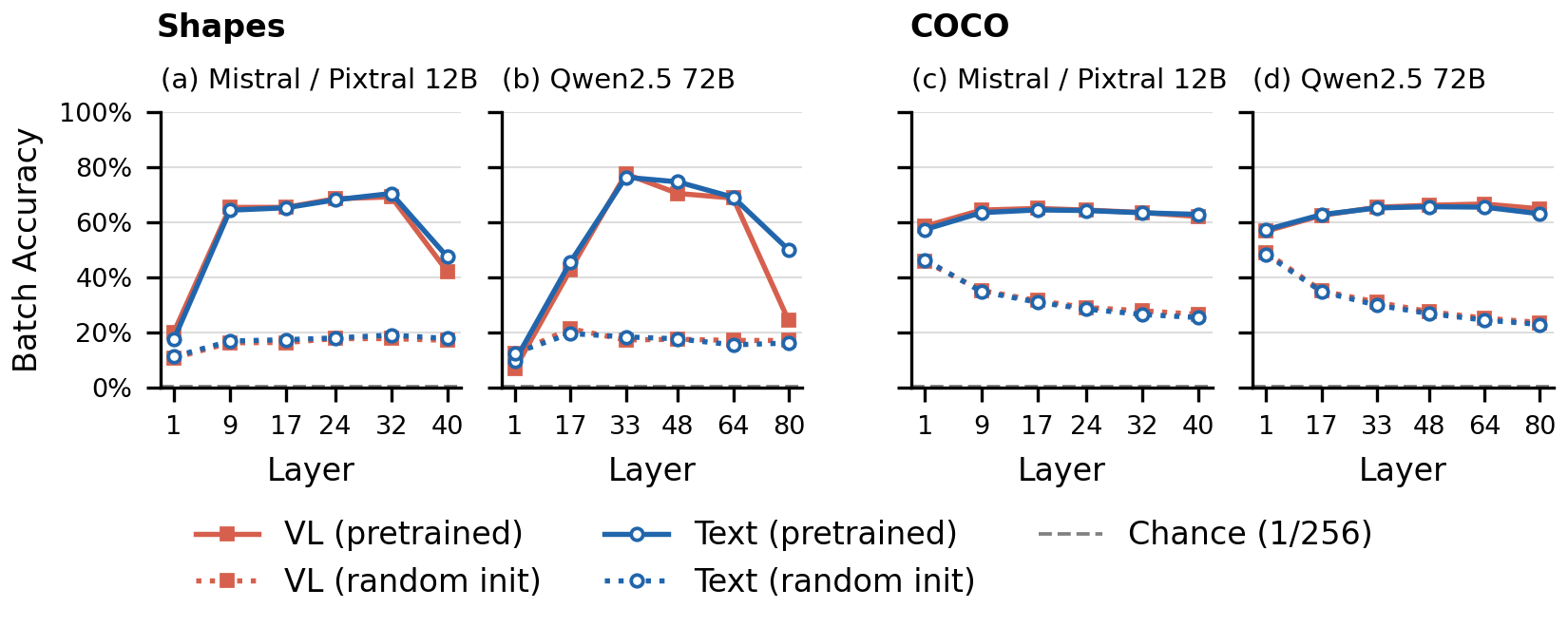}
    \caption{We observe the same trend as in Figure~\ref{fig:contrastive-probe} i.e. the text embeddings from the text-only and the VL models show similar propensity to align with the corresponding image embeddings. The absolute numbers are higher than seen in Figure~\ref{fig:contrastive-probe} because siglip image embeddings are pre-trained with text supervision and it is not a ``pure'' image encoder.}
    \label{fig:contrastive-probe_siglip}
\end{figure}

\subsection{LM Evaluations}\label{app:lm-eval}

These evaluations were run using the LM-Eval harness~\citep{eval-harness}. We run the commonsense 0 shot setting, which consists of evaluating on Multiple-Choice questions on Arc challenge~\citep{clark2018thinksolvedquestionanswering}, Hellaswag~\citep{zellers2019hellaswagmachinereallyfinish}, PIQA~\citep{bisk2019piqareasoningphysicalcommonsense}, and Winogrande~\cite{sakaguchi2019winograndeadversarialwinogradschema}. We also run a 5-shot ICL setting on MMLU~\citep{hendrycks2021measuringmassivemultitasklanguage}. See results in Table~\ref{tab:lmeval}.

\begin{table}[h]
\centering
\begin{tabular}{lccc}
\toprule
& VL & Text & Distilled \\
\midrule
Commonsense 0 Shot (Accuracy) & 60.33\% & 59.39\% & 59.10\% \\
MMLU (macro avg)          & 72.49\% & 74.66\% & 70.14\% \\
\bottomrule
\end{tabular}
\caption{There is minimal to no degradation from the initialized VL checkpoint due to the distillation.}\label{tab:lmeval}
\end{table}

\subsection{How often do models get the total number of shapes right?}\label{app:expt_1_copout}

\begin{figure}[htbp]
    \centering
    \includegraphics[width=\textwidth]{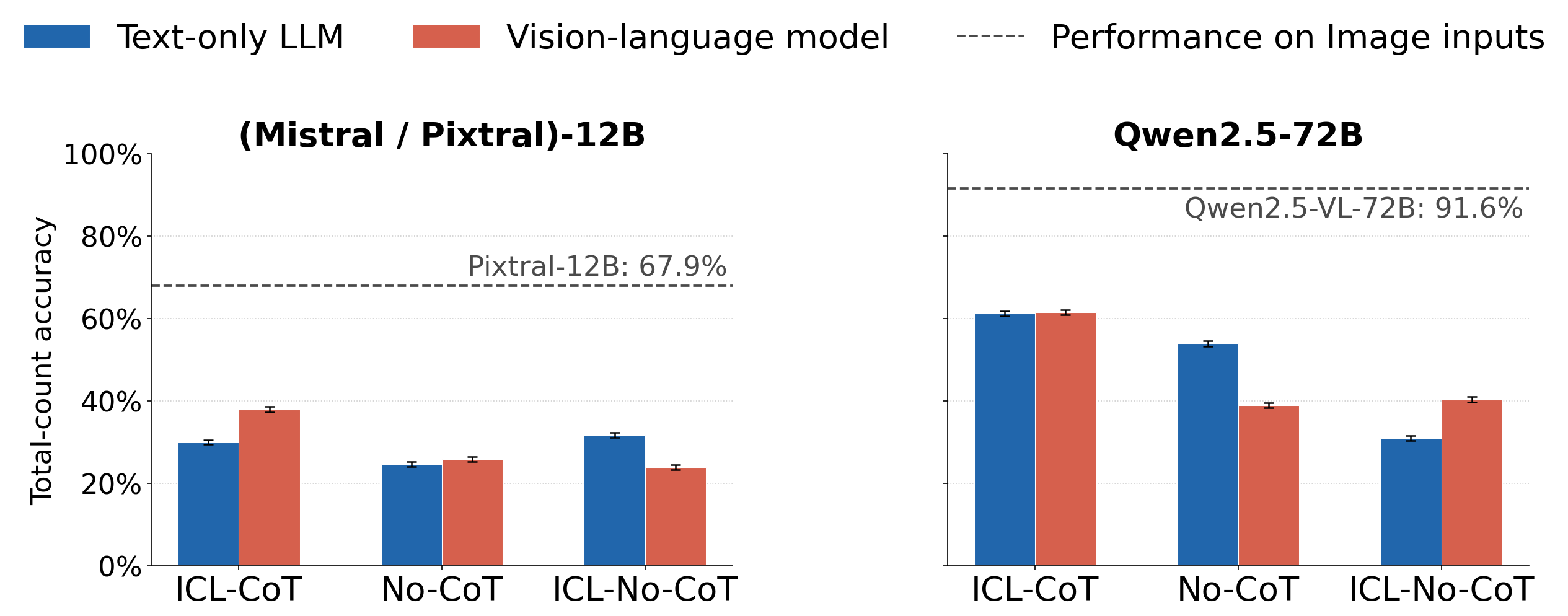}
    \caption{The models perform much better at the easier sub-task of getting the total number of shapes in a sample.}
    \label{fig:behavioral_total_shapes_count}
\end{figure}

We note that models can get the total number of shapes right without identifying them by performing a closed-loop analysis as previously mentioned in Section~\ref{sec:behavioral-experiment}. They can look for cycles in the vertices and group them to deduce shapes. For our largest model (Qwen 2.5 72B), the COT setting really helps in getting the shape count variant right. See Figure~\ref{fig:behavioral_total_shapes_count}.

\subsection{Probe Results on Distilled Artifacts}

\begin{figure}[htbp]
    \centering
    \begin{subfigure}[b]{0.48\textwidth}
        \centering
        \includegraphics[width=\textwidth]{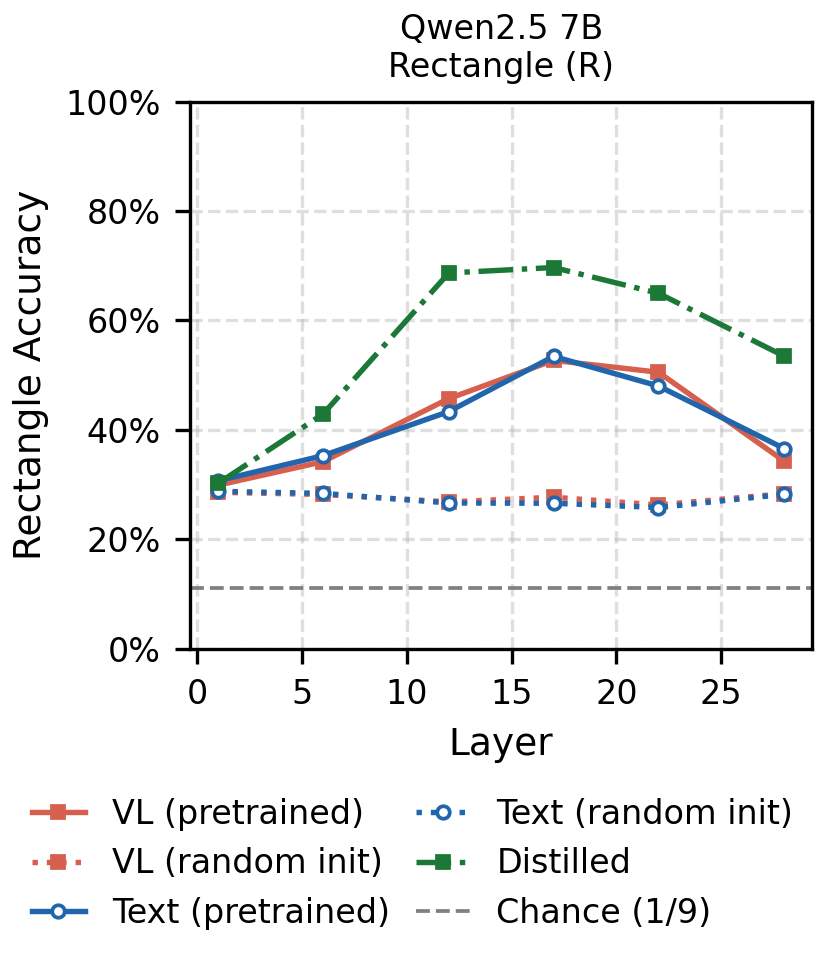}
        \caption{Linear probe on rectangles}
        \label{fig:sub1}
    \end{subfigure}
    \hfill
    \begin{subfigure}[b]{0.48\textwidth}
        \centering
        \includegraphics[width=\textwidth]{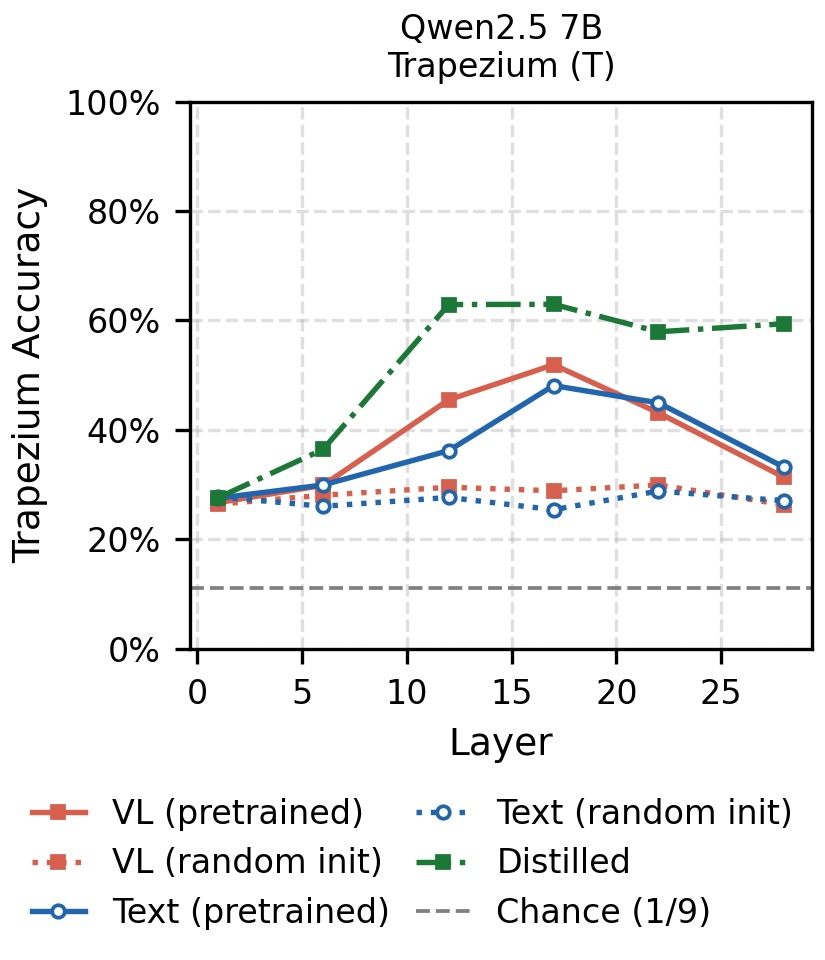}
        \caption{Linear probe on trapeziums}
        \label{fig:sub2}
    \end{subfigure}
    \caption{The distilled model representations are better across layers.}
    \label{fig:distilled_linear_probe}
\end{figure}

\begin{figure}[htbp]
    \centering
    \includegraphics[width=0.8\textwidth]{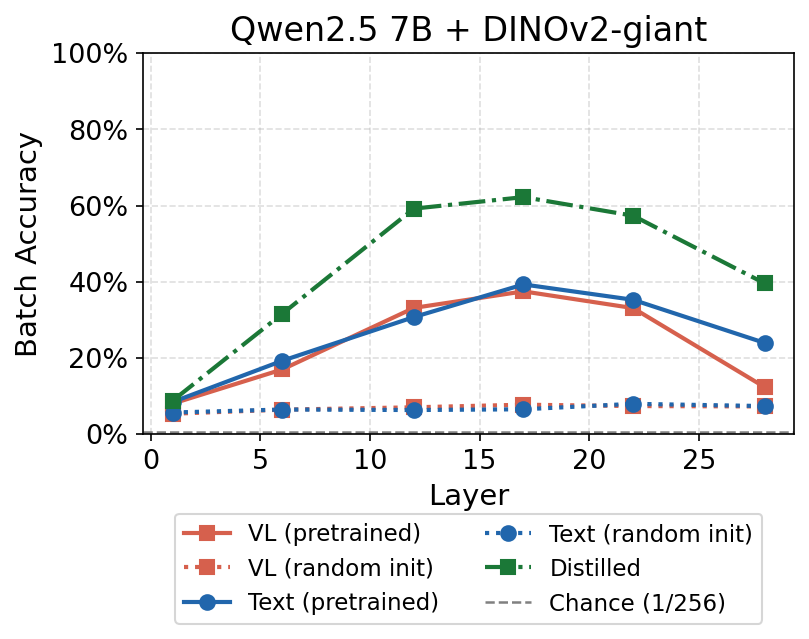}
    \caption{The distilled model representations are better across layers.}
    \label{fig:distilled_contrastive_probe}
\end{figure}

We run the probe experiments from Section~\ref{sec:representation_expt} on our distilled artifacts from Section~\ref{sec:distil_section}. The distillation objective, in some sense, is a literalisation of our strong multimodality principle as an optimization objective. Therefore, if the probes are measuring the right proxy, we should see a strong difference in trends for the distilled model on those experiments compared to the release models. We present these results in Figure~\ref{fig:distilled_linear_probe} and Figure~\ref{fig:distilled_contrastive_probe} and see that the distilled model's representations not only perform better on the probing task but also lend themselves better for the contrastive (retrieval) task against an unrelated image encoder. This is consistent with the probes tracking the property the distillation targets, though in-distribution fine-tuning may also raise probe accuracy.

\end{document}